# 1st Place Solution of Egocentric 3D Hand Pose Estimation Challenge 2023 Technical Report: A Concise Pipeline for Egocentric Hand Pose Reconstruction

Zhishan Zhou[1], Zhi Lv[1], Shihao Zhou, Minqiang Zou,
Tong Wu, Mochen Yu, Yao Tang, Jiajun Liang[*]
Jiiov Technology
{zhishan.zhou, zhi.lv, shihao.zhou, minqiang.zou, tong.wu
mochen.yu, yao.tang, jiajun.liang}@jiiov.com

## Abstract

*This report introduce our work on **Egocentric 3D Hand Pose Estimation** workshop. Using AssemblyHands[9], this challenge focuses on egocentric 3D hand pose estimation from a single-view image. In the competition, we adopt ViT based backbones and a simple regressor for 3D keypoints prediction, which provides strong model baselines. We noticed that Hand-objects occlusions and self-occlusions lead to performance degradation, thus proposed a non-model method to merge multi-view results in the post-process stage. Moreover, We utilized test time augmentation and model ensemble to make further improvement. We also found that public dataset and rational preprocess are beneficial. Our method achieved **12.21mm** MPJPE on test dataset, achieve **the first place** in Egocentric 3D Hand Pose Estimation challenge.*

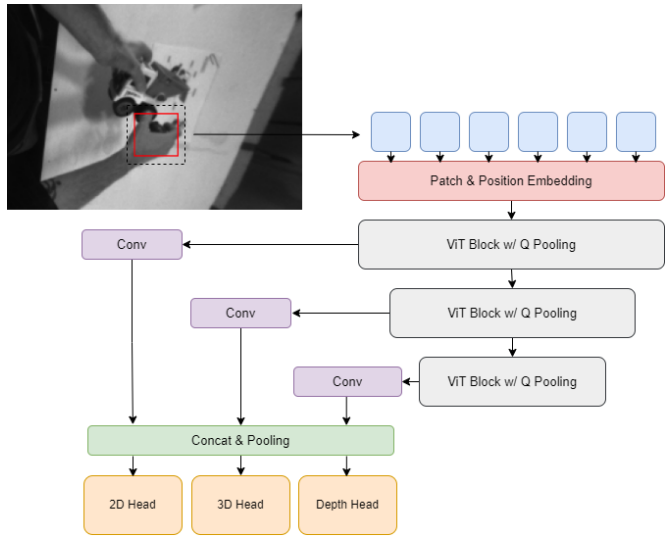

Figure 1. The input image is cropped according to the hand bounding box, feeding to a vision transformer to extract features. MLP headers are used to regress 2D keypoints, root-relative 3D keypoints and root depth.

## 1. Introduction

3D hand pose estimation plays an important role in gesture interaction applications. Although the paid efforts, It remains as a challenging research task due to high degree of freedom and vulnerability in occlusion scenarios.

Existing approaches can be roughly categorized into two types: optimizing the accuracy of visible points and enhancing the stability of occluded points. For the former, Typical works proposed to use multiple stage models to precisely locate keypoints through iterative refinement[8], or utilizing prior structural information, such as hand skeleton, to eliminate ambiguity in gesture modeling[4]. For the latter, Some methods employed a multi-view camera system, leveraging complementary features of each viewpoints to mitigate the negative effects of occlusion. [5] incorporated historical frame information to estimate the occluded keypoints in the current frame, to maintain temporal consistency. Additionally, Ziani et al. [11] leveraged unsupervised hands data for pre-training, improved the performance with out synthetic data or specialized architectures, shows the potential of pre-trained models.

In this work, we provide useful approaches to improve accuracy for 3D keypoints prediction. We validated that a robust feature extraction backbone and a straightforward 3D skeleton decoder is sufficient to achieve high accuracy. In addressing occluded points, we propose a non-trivial approach by calculating the similarity between monocular results, multi-view results, and history frames, ultimately merging the high quality outputs. Our experiment also shows the necessity of undistortion on fish-eye camera and the strong impact of the cropped hands area. More details

---

1 equanly contribution, * corresponding author will be explained in the following section.

## 2. Method

**Datasets** Data diversity and amount is important for deep learning models especially large models, so we collect several public datasets for hand pose estimation which is allowed by competition rules. AssemblyHands[9] is a multi-views egocentric-image dataset containing videos of participants assembling and disassembling take-apart toys. Freihand[3] is a single-view color image dataset collected with green screen background which is then replaced with different various background images. DexYCB[1] captures videos of hand grasping objects. CompHand[2] is a synthesised dataset with multiple pose and viewpoints. We find that additional data lead to about 0.3mm MPJPE(mean per-joint position error) improvement.

**Model Structure** Vision Transformers have recently shown powerful performance in computer vision problems. We choose a recently proposed fast and powerful vision transformer architecture, Hiera[10], pretrained with MAE[6] as our backbone to extract features for hand pose estimation problem. We use simple MLP head for regressing 2D keypoints, root-relative 3D keypoints and root depth. A multi-level feature fusion operation which concatenates the features of different layers is adopted to better extract hand feature with different scale. The overview of our model is shown in Figure1

**Preprocess** We observed that hands near the edge of the rectified image which is obtained by an undistortion operation on the fisheye image are stretched too much. Such a deformation may be hard for the model to learn the depth information. The small proportion of such data make the learning efficiency even worse. So we conduct a warp perspective operation to make the hands near the edge less stretched. The transformation matrix is calculated by virtually rotating the camera towards the hand. We also set an enlarge scale to the hand bounding box according to hand scale and two hand distance which we will discuss in the next section.

**Augmentation** We randomly flip the image horizontally to get more training samples. We also apply channel noise by multiplying each channel with a random scale between 0.6 and 1.4 and brightness contrast adjustment to the image. Additionally, we randomly mask some part of the image to simulate occlusion.

**Postprocess** Since the outputs of the model are located in each camera coordinate, we apply a multi-view merge strategy to get the final world coordinate results. Specifically, we compute the MPJPE with each other view and select two results of views which have the lowest MPJPE. If the MPJPE is lower than a threshold, the mean of the two results is calculated as final result. Otherwise we choose the result which has a lower PA-MPJPE with the result of previous frame. After the multi-view merging operation, we apply an offline smooth filter on each video.

## 3. Experiments

We use the datasets mentioned in Section 2. we use 8 2080Ti GPUs (16 images/GPU) therefore 128 images were utilized per minibatch. The model was trained for 300/600 epoch. We use AdamW[7] optimizer with an initial learning rate of 1e-4 scheduled by cosine scheduler and weight decay 1e-2. Despite commonly used augmentations, we apply random mask to the image with a probability of 0.5, making it robust to occlusion. We notice that two-hands interaction are quite rare in test set, so no particular optimization was made.

We first train three Hiera large models and one convnext model. For each model, we aggregate the predicted multi-view outcomes into egocentric results and smooth the fused results by applying a Savitzky-Golay filter. Secondly, we merge multi-scale crop result, specifically, we use different crop scale for different videos. If there are many small hands in the video or lots of situations where both hands are covering each other, do not expand the crop area of hand. Otherwise, expand the crop area of hand to get better performance. Finally, we ensemble the result of three Hiera large models, and merge ConvNext results into that of Hiera's. When merging, hiera is default to be set to a larger weight because of it's better performance. When there is a significant gap between two results, we increase the weight of ConvNext.

| ID | method | MPJPE |
|---|---|---|
| 1 | hiera + multi-view + smooth | 13.6 |
| 2 | ID1 + muti-scale crop | 12.9 |
| 3 | ID2 + multi-model fusion | 12.7 |
| 4 | ID3 + convNext fusion | 12.2 |

Table 1. Quantitative results on Assembly101 test set.

The final result was shown in Table 1. with applied multi-view aggregation and smoothing, the best single model performance is 13.6mm. Taking TTA into consideration, multi-scale cropping improve a large margin to 12.7mm. As to ensemble, hiera model merging raise the performance slightly by 0.2mm, ConvNext merging brings further improvement from 12.7mm to 12.2mm.

## 4. Conclusion

We conducted an in-depth study on 3D keypoint tasks, proposing several effective methods from aspects of data collection, data processing, model structure, and fusion methods. Our approach achieved promising results and obtained the best performance in Egocentric 3D Hand Pose Estimation challenge.